\documentclass[letterpaper, 10 pt, conference]{ieeeconf}  

\usepackage{color}
\usepackage[latin1]{inputenc}
\usepackage{multirow}
\usepackage{graphicx}
\usepackage{epstopdf}
\usepackage{bbm}     
\usepackage{amsmath, amssymb,amstext}
\usepackage{bm}
\usepackage{amsmath}
\usepackage{amsfonts}
\usepackage{booktabs} 
\usepackage{threeparttable} 
\usepackage{tabularx}
\usepackage{subfigure}

\usepackage{float}
\usepackage{url}
\usepackage{cuted}
\usepackage[ruled,linesnumbered,titlenumbered]{algorithm2e}
\usepackage[linkcolor=black,citecolor=black,urlcolor=black,colorlinks=true]{hyperref}

\IEEEoverridecommandlockouts                              

\title{\LARGE \bf Distributed Motion Coordination Using Convex Feasible Set Based\\ Model Predictive Control}

\author{Hongyu Zhou$^{1}$ and Changliu Liu$^{2}$
\thanks{$^{1}$Hongyu Zhou is with the Department of Marine Technology, Noweigian University of Science and Technology, NO-7491 Trondheim, Norway
        {\tt\small hongyuz@alumni.ntnu.no}}%
\thanks{$^{2}$Changliu Liu is with the The Robotics Institute,
	Carnegie Mellon University, 5000 Forbes Avenue, Pittsburgh, PA, 15213, USA
        {\tt\small cliu6@andrew.cmu.edu}}%
}

\begin{document}

\maketitle
\thispagestyle{empty}
\pagestyle{empty}

\begin{abstract}
	

The implementation of optimization-based motion coordination approaches in real world multi-agent systems remains challenging due to their high computational complexity and potential deadlocks. This paper presents a distributed model predictive control (MPC) approach based on convex feasible set (CFS) algorithm for multi-vehicle motion coordination in autonomous driving. By using CFS to convexify the collision avoidance constraints, collision-free trajectories can be computed in real time. We analyze the potential deadlocks and show that a deadlock can be resolved by changing vehicles' desired speeds. The MPC structure ensures that our algorithm is robust to low-level tracking errors. The proposed distributed method has been tested in multiple challenging multi-vehicle environments, including unstructured road, intersection, crossing, platoon formation, merging, and overtaking scenarios. The numerical results and comparison with other approaches (including a centralized MPC and reciprocal velocity obstacles) show that the proposed method is computationally efficient and robust, and avoids deadlocks.

\end{abstract}


\section{Introduction}
The research on autonomous vehicles receives increasing attentions in the robotics community, due to its potential to improve the mobility and efficiency of the transportation systems. However, it remains challenging for autonomous vehicles to properly interact with other road participants. This paper considers the interaction among a group of autonomous vehicles which are assumed to be connected, i.e., able to share low-bandwidth information. To realize the benefits of autonomous vehicles, it is important to develop safe and efficient methods to control and coordinate these connected autonomous vehicles in a distributed fashion. 

Classical methods for multi-agent coordination include potential field \cite{wachter2008potential}, \cite{tanner2005towards}, reciprocal velocity obstacles (RVO) \cite{van2008reciprocal}, and scheduling designs \cite{liu2017distributed}, \cite{azimi2013reliable}. However, they do not explicitly consider the interaction among agents. Methods like cell decomposition \cite{guo2002distributed}, \cite{bennewitz2001optimizing} and roadmap approaches \cite{svestka1995coordinated}, \cite{wurm2008coordinate} reduce the continuous motion planning problem to a discrete graph search problem, and therefore, usually result in non-smooth trajectories and suboptimal solutions. 

Optimization-based methods which plan in continuous space can generate smoother trajectories. Besides, it is able to take into account the interaction among agents by formulating constraints properly. \cite{firoozi2020distributed} formulates collision avoidance as a dual optimization problem and proposes a bi-level distributed MPC scheme. \cite{keviczky2007decentralized} uses invariant-set theory and mix-integer linear programming (MILP).  \cite{rey2018fully} and \cite{ferranti2018coordination} rely on the alternating direction method of multipliers (ADMM). They are not computationally efficient for real time applications and scale poorly with the number of vehicles, since the optimization problem is usually nonlinear and non-convex.

Convex feasible set (CFS) algorithm is an optimization algorithm for real time motion planning \cite{liu2017real}, \cite{liu2017convex}, \cite{liu2018convex}. It handles optimization problems with convex objective function and non-convex constraints. Like sequential quadratic programming (SQP) \cite{johansen2004constrained}, \cite{schulman2013finding}, \cite{ziegler2014trajectory}, CFS algorithm approximates the original problem as a sequence of convex sub-problems. By exploiting the unique geometric structure of motion planning problems, it is an order of magnitude faster than SQP. Based on CFS algorithm, a fast MPC-based motion planner \cite{chen2018foad} and a centralized multi-vehicle planner (MCCFS algorithm) \cite{MCCFS2020} have been proposed. While CFS algorithm makes the real-time optimization-based planning possible, it is still challenging to handle deadlocks and tracking errors.


In this paper, we propose a CFS-based distributed MPC design, called CFS-DMPC, for efficient, safe, and coordinated multi-vehicle motion planning in autonomous driving. We explicitly exploit the structure of the coordination problem to formulate the distributed approach, with the assumption that vehicle to vehicle (V2V) communication is available. Our method does not rely on MILP, ADMM, or a high-level decision maker for scheduling. Instead, we leverage the CFS algorithm to convexify the collision avoidance constraints and make the distributed MPC approach more efficient. This also avoids solving an additional optimization problem for collision avoidance as in \cite{firoozi2020distributed}. We analyze the features of the deadlock situation and propose a solution by changing vehicles' desired speeds. We validate our method through simulation, showing its robustness to tracking errors and comparing its performance with a centralized implementation, a scheduling design, and RVO. The source code is available at \textit{\href{https://github.com/intelligent-control-lab/Auto_Vehicle_Simulator}{https://github.com/intelligent-control-lab/Auto\_Vehicle\_Simulator}}.

The remainder of the paper is organized as follows. Section \ref{Ch2} introduces multi-vehicle motion coordination problem and CFS algorithm. Section \ref{Ch3} formulates the CFS-DMPC design. Section \ref{Ch4} presents the numerical results. Section \ref{Ch5} concludes the paper with directions for future work.

\section{Problem Formulation}
\label{Ch2}

\subsection{Multi-vehicle Motion Coordination}
The multi-vehicle motion coordination can be formulated as a centralized MPC which computes the collision-free trajectories for all vehicles simultaneously. 

\subsubsection{Configuration Space}
For an autonomous vehicle, the configuration space is its 2D position, denoted as $x \in \mathbb{R}^2$. Note that the heading is ignored here and will be taken care of in the low-level tracking controller. Then the trajectory of the $i$th vehicle can be denoted as $ \mathbf{x}_i =  \left[ x_i^1 ; x_i^2 ; ... ; x_i^H\right] \in \mathbb{R}^{2H}  $, where $H$ is the planning horizon, $i \in \mathcal{V}$ is the vehicle index, and $\mathcal{V} := \{1,2,...,N \}$ is the set of $N$ vehicles. 


\subsubsection{Objective Function}
We formulate the objective function $ J_i(\mathbf{x}_i,s_i) $ for the $i$th vehicle to be quadratic. It is defined as $J_i(\mathbf{x}_i,s_i) = J_i^o(\mathbf{x}_i) + J_i^a(\mathbf{x}_i) + J_i^s(s_i)$. The three terms are explained as follows.

$J_i^o(\mathbf{x}_i)$ penalizes the difference between the planned trajectory $\mathbf{x}_i$ and the reference trajectory $\mathbf{x}_i^{ref} \in \mathbb{R}^{2H}$. $\mathbf{x}_i^{ref}$ can be the centerline of the target lane for each vehicle. $J_i^o(\mathbf{x}_i)$ is given by 
\begin{equation}
\begin{aligned}
J_i^o(\mathbf{x}_i) &=  \frac{1}{2}c_o (\mathbf{x}_i-\mathbf{x}_i^{ref})^\top(\mathbf{x}_i-\mathbf{x}_i^{ref}) \\
&= \frac{1}{2}c_o\mathbf{x}_i^\top\mathbf{x}_i - c_o\mathbf{x}_i^\top\mathbf{x}_i^{ref} + \frac{1}{2}c_o(\mathbf{x}_i^{ref}) ^\top\mathbf{x}_i^{ref},
\end{aligned}
\end{equation}
where $c_o$ is a weighting parameter. Note that $\frac{1}{2}c_o(\mathbf{x}_i^{ref}) ^\top\mathbf{x}_i^{ref}$ can be neglected since it is constant.

$J_i^a(\mathbf{x}_i)$ penalizes the norm of net acceleration, defined as 
\begin{equation}
J_i^a(\mathbf{x}_i) =  \frac{1}{2}c_a (\mathbf{A}_i\mathbf{x}_i)^\top(\mathbf{A}_i\mathbf{x}_i) = \frac{1}{2}c_a \mathbf{x}_i^\top \mathbf{A}_i^\top\mathbf{A}_i\mathbf{x}_i,
\end{equation}
where $c_a$ is a weighting parameter and $\mathbf{A}_i$ is a linear operator that maps the trajectory $\mathbf{x}_i$ to the accelerations along the trajectory. The mapping depends on the sampling time $T_s$.	The effect of penalizing $J_i^a(\mathbf{x}_i)$ is to make the planned trajectory smooth, such that it can be easily tracked.


$ J_i^s(s_i) = c_s \left\| s_i \right\|^2 $ penalizes the magnitude of the slack variable $s_i \in \mathbb{R}^2$ which will be introduced later, with $c_s$ as a weighting parameter.

\subsubsection{Constraints}
In this paper, the collision avoidance constraints force each vehicle pair $(i,j)$, where $i,j \in \mathcal{V}$ and $i\neq j$, to maintain a safety distance at every time step:
\begin{equation}
\phi({x}_i^h,{x}_j^h) = d({x}_i^h,{x}_j^h) - d_{min} \geq 0, \forall h \in \{1,2,...,H\},
\end{equation}
where  $ d({x}_i^h,{x}_j^h) $ is the distance between ${x}_i^h$ and ${x}_j^h$, and $d_{min}$ is the safety margin between two vehicles, which can take into account the model uncertainty and measurement errors. Note that the collision avoidance constraints are non-convex.

In addition to the safety constraints, we want the planned trajectories to start from vehicles' current positions. Inspired by \cite{chen2018foad}, we introduce a slack variable to this condition to generate a smoother planned trajectory. Hence, we modify this constraint as ${x}_i^1 = {x}_i^{c} + s_i$, where ${x}_i^c$ represents the current position of the $i$th vehicle.

The resulting centralized MPC is summarized as
\begin{subequations}
	\begin{align}
	\mathop{\min}_{\mathbf{x}_i,s_i} \quad &  \sum_{i=1}^{N}  J_i(\mathbf{x}_i,s_i)  & \label{Eq_CMPC_obj} \\
	\textrm{s.t.} \quad & \phi({x}_i^h,{x}_j^h) = d({x}_i^h,{x}_j^h) - d_{min} \geq 0, & \label{Eq_CMPC_constraint}  \\
	& {x}_i^1 = {x}_i^{c} + s_i,
	\end{align}
	$\quad\quad\quad\quad\ \  \forall i \in \mathcal{V}, j \in \mathcal{V}\backslash\{i\}, h \in \{1,2,...,H\}.$
	\label{Eq_CMPC_opt}
\end{subequations}

\subsection{Convex Feasible Set (CFS) Algorithm}
The CFS algorithm \cite{liu2018convex} solves non-convex optimization problems in the following form:
\begin{equation}
\begin{array}{rrclcl}
\displaystyle   \mathbf{x}^* = \mathop{\arg\min}_{\mathbf{x} \in \Gamma }  J(\mathbf{x}),
\end{array}
\label{Eq_CFS_opt}
\end{equation}
where $\mathbf{x}$ is the state variable, $J(\mathbf{x})$ is a smooth and convex objective function, $\Gamma$ is the state space  constraint which can be non-convex, and $\mathbf{x}^*$ is the optimal solution. Problem \eqref{Eq_CMPC_opt} can be written in the form of \eqref{Eq_CFS_opt}. There are three steps for CFS algorithm to solve the optimization problem \eqref{Eq_CFS_opt}.

\textit{Step 1}: Initialize the state variable $\mathbf{x}^{(0)}$.

\textit{Step 2}: Using the state variable of the last iteration $\mathbf{x}^{(k)}$, calculate the convex feasible set $\mathcal{F}(\mathbf{x}^{(k)}) \subset \Gamma$. In practice, 5$\Gamma$ is an intersection of multiple constraints, i.e.,  $\Gamma = \cap_i\Gamma_i$, where $\Gamma_i = \{ \mathbf{x} : \phi_i \geq 0 \}$ is the space outside of the $i$th obstacle, and $\phi_i$ is the signed distance function to the $i$th obstacle. When the $i$th obstacle is convex, the corresponding convex feasible set can be computed as
\begin{equation}
\small
\mathcal{F}_i(\mathbf{x}^{(k)}) = \{ \mathbf{x}: \phi_i(\mathbf{x}^{(k)}) +  \nabla \phi_i(\mathbf{x}^{(k)})(\mathbf{x}-\mathbf{x}^{(k)}) \geq 0    \},
\label{Eq_CFS}
\end{equation}
where $\nabla$ is the gradient operator. At a point where $\phi_i$ is not differentiable, $\nabla \phi_i$ is a sub-gradient such that the convex feasible set $\mathcal{F}:=\cap_i\mathcal{F}_i$ always includes the steepest descent direction of $J(\mathbf{x})$ in the set $\Gamma$ \cite{liu2017convex}.

\textit{Step 3}: Given the convex feasible set $\mathcal{F}(\mathbf{x}^{(k)})$, obtain the new solution $\mathbf{x}^{(k+1)}$ by solving 
\begin{equation}
\begin{array}{rrclcl}
\displaystyle   \mathbf{x}^{(k+1)} = \mathop{\arg\min}_{\mathbf{x} \in \mathcal{F}(\mathbf{x}^{(k)}) }  J(\mathbf{x}).
\end{array}
\label{Eq_CFS_cvxopt}
\end{equation}

The CFS algorithm starts with step 1, then applies steps 2 and 3 iteratively. It is proved in \cite{liu2018convex} that the sequence $\mathbf{x}^{(k)}$ generated by step 3 will converge to a local optimum $\mathbf{x}^*$ of the optimization problem \eqref{Eq_CFS_opt}.

A centralized planner called MCCFS \cite{MCCFS2020} has applied CFS algorithm to solve the multi-vechicle coordination problem \eqref{Eq_CMPC_opt}. However, it has limited scalability with respect to the number of vehicles. This paper intends to develop a CFS-based distributed approach to efficiently solve multi-vehicle coordination problems.

\section{Distributed Motion Coordination}
\label{Ch3}

A distributed approach for motion coordination must address three challenges. First, it needs to break the coupling among vehicles in the problem~\eqref{Eq_CMPC_opt} and allow each vehicle to efficiently plan in parallel. Second, the approach should be deadlock-free. Finally, the approach should be robust to tracking errors.

In this section, we present a distributed approach which allows each vehicle to solve a sub-problem in parallel. Based on CFS algorithm, the non-convex sub-problem is transformed into a QP problem, which can be solved efficiently. We analyze potential deadlocks and propose a general deadlock resolution strategy. We utilize MPC to ensure our method is robust to low-level tracking errors.

The overall algorithm for vehicle $i$ is presented in Algorithm \ref{alg_DCFS}. The trajectory is replanned every $T_r$ seconds. First, vehicle $i$ communicates with all surrounding vehicles (line 3). Line 4 implements the proposed deadlock resolution strategy. If vehicle $i$ detects a deadlock (according to the criteria in \eqref{Eq_Deadlock}), it changes its desired speed. Then if the deadlock is resolved and vehicle $i$ reaches its reference trajectory, it is allowed to move at its original speed. Finally, $\mathbf{x}_i^{ref}$ should be modified according to its current position ${x}_i^{c}$ and the desired speed. The optimal trajectory for vehicle $i$ can be obtained by solving a sub-problem (lines 5-7). Note that this optimal solution is used to initialize $\mathbf{x}_i^{(0)}$ in the next planning.
\begin{algorithm}[t]
	\small
	\caption{The CFS-DMPC design for vehicle $i$}
	\label{alg_DCFS}
	\SetKwInput{KwPara}{Parameter}
	\KwIn{ ${x}_i^{c}$, $\mathbf{x}_j$, $\forall j \in \mathcal{V} \backslash\{i\}$ }
	\KwPara{ $c_o$, $c_a$, $c_s$, $T_r$, $T_s$, $H$, $l$, $w$, $r$, $n$, $\epsilon_1$, $\epsilon_2$}
	\KwOut{ $\mathbf{x}_i$ }  
	\BlankLine
	Initialize $\mathbf{x}_i$, $\mathbf{x}_i^{ref}$; \\
	

	\For{$t = 0, T_r, 2T_r, ... , \infty$}{
		Communication with vehicle $j$, $\forall j \in \mathcal{V} \backslash\{i\}$ : send $\mathbf{x}_i$ and receive $\mathbf{x}_j$; \\
		Check deadlocks and change the desired speed accordingly; \\	
		Modify $\mathbf{x}_i^{ref}$	according to ${x}_i^{c}$ and the desired speed; \\
		Initialize $\mathbf{x}_i^{(0)}$ with $\mathbf{x}_i$ from the previous planning;\\ 
		Solve optimization problem \eqref{Eq_DCFS_opt} for $\mathbf{x}_i$.
	}
	
\end{algorithm}

\subsection{CFS-based Distributed MPC Formulation}
Vehicles are coupled in problem \eqref{Eq_CMPC_opt} due to collision avoidance constraints \eqref{Eq_CMPC_constraint}. To break the coupling and present a distributed approach, we assume vehicles are able to send and receive the planned trajectories. The vehicle $i$ uses received trajectories from surrounding vehicles for motion planning, denoted by $\bar{\mathbf{x}}_j = \left[ \bar{x}_j^1 ; \bar{x}_j^2 ; ... ; \bar{x}_j^H\right]$ for all $j\in \mathcal{V}\backslash\{i\}$. The collision avoidance constraints for vehicle $i$ now become $\phi({x}_i^h,\bar{x}_j^h) = d({x}_i^h,\bar{x}_j^h) - d_{min} \geq 0$ for all $j$ and $h$. 
This strategy removes coupling terms in the centralized MPC. 



\begin{figure}
	\centering
	\includegraphics[width=0.4\textwidth]{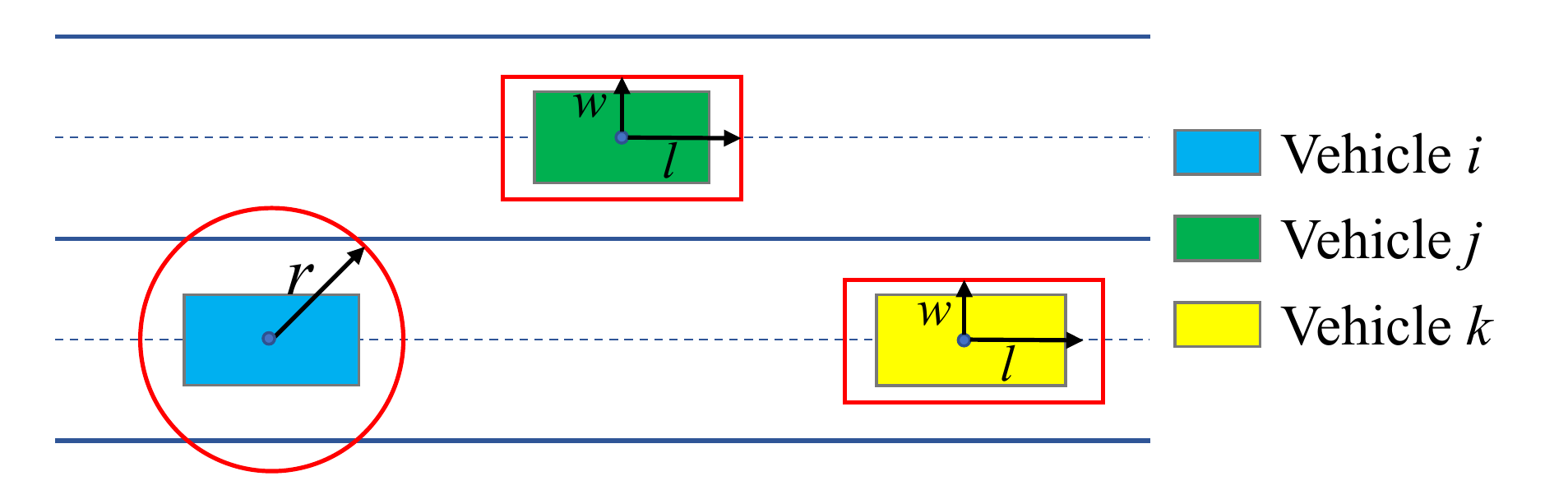}
	\vspace{-4mm}	
	\caption{Vehicles' representation for collision avoidance from the perspective of vehicle $i$.}
	\label{fig_VehRep}
	\vspace{-20pt}
\end{figure}

To leverage CFS algorithm, the geometric representation for vehicles from the perspective of vehicle $i$ is defined as shown in Fig.~\ref{fig_VehRep}. Vehicle $i$ itself is represented by a circle with radius $r$, and each surrounding vehicle is represented by a rectangle with length $2l$ and width $2w$. We use $\mathcal{O}$ to denote the rectangle and further modify the collision avoidance constraints as $\phi({x}_i^h,\bar{\mathcal{O}}_j^h) = d({x}_i^h,\bar{\mathcal{O}}_j^h) - r \geq 0$, where $d({x}_i^h,\bar{\mathcal{O}}_j^h)$ is the signed distance function from ${x}_i^h$ to $\bar{\mathcal{O}}_j^h$ \cite{liu2018convex}. Since the rectangular shape is convex, \eqref{Eq_CFS} can be used to find the convex feasible set, given by
\begin{equation}
\phi_{i,j}^{h,k} +  \nabla \phi_{i,j}^{h,k} \  ({x}_i^{h}-{x}_i^{h(k)}) \geq 0,
\label{Eq_DCFS_constraint_formulation}
\end{equation}
where $\phi_{i,j}^{h,k}:=\phi({x}_i^{h(k)},\bar{\mathcal{O}}_j^h)$ and $k$ is the iteration number.

Replacing constraint \eqref{Eq_CMPC_constraint} by \eqref{Eq_DCFS_constraint_formulation}, the CFS-DMPC design for vehicle $i$ is given by
\begin{subequations}
	\begin{align}
	\mathop{\min}_{\mathbf{x}_i,s_i} \quad &  J_i(\mathbf{x}_i,s_i) & \label{Eq_DCFS_obj} \\
	\textrm{s.t.} \quad & \phi_{i,j}^{h,k} +  \nabla \phi_{i,j}^{h,k} \  ({x}_i^{h}-{x}_i^{h(k)}) \geq 0 & \label{Eq_DCFS_constraint}  \\
	& {x}_i^1 = {x}_i^{c} + s_i,
	\end{align}
	$\quad\quad\quad\quad\quad \forall j \in \mathcal{V}\backslash\{i\}, h \in \{1,2,...,H\}. $
	\label{Eq_DCFS_opt}
\end{subequations}

Note that problem \eqref{Eq_DCFS_opt} is indeed a QP problem. Since the solution in any iteration is guaranteed to satisfy the original collision avoidance constraint, we can safely stop the iteration before convergence. For the sake of computation efficiency, we only solve it for one iteration during MPC replanning in the following discussion.

\subsection{Deadlock Breaking}
\begin{figure}
	\centering
	\subfigure[Two vehicles in a deadlock situation.]{\includegraphics[width=0.23	\textwidth]{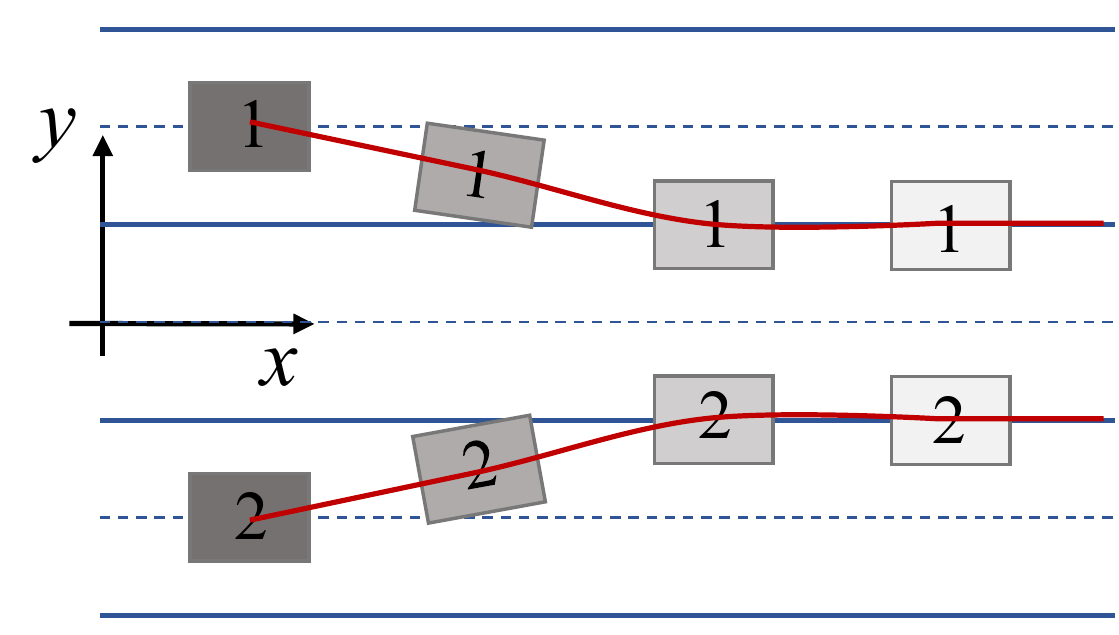}\label{fig_Deadlock}}	
	\subfigure[Forming a platoon with the proposed deadlock resolution. ]{\includegraphics[width=0.23\textwidth]{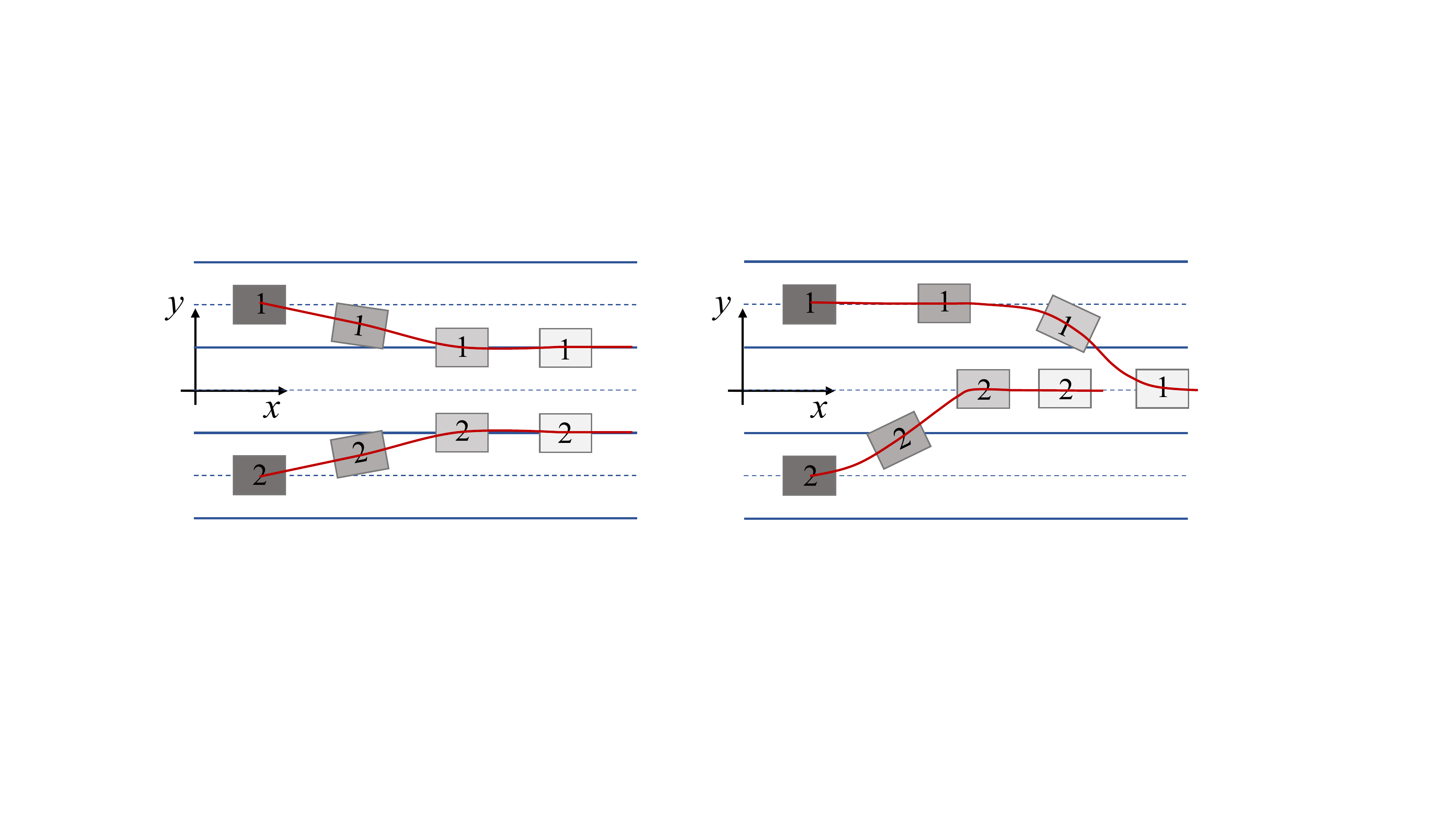}\label{fig_DeadlockBreaking}}
	\vspace{-3mm}	
	\caption{Demonstration of deadlock breaking. The lighter color represents the future time steps and the red line is the planned trajectory.}
	\label{fig_DeadlockDemo}
\end{figure}

A deadlock occurs when multiple vehicles have the same or symmetric reference trajectories so that none of the vehicles can find a collision free trajectory that converges to the reference. An example of two vehicles stuck in a deadlock is shown in Fig. \ref{fig_Deadlock}. Vehicles~1 and 2 have symmetric positions about $x$-axis and same speed. They have the same reference trajectory, which is a sequence of points on $y=0$. Instead of forming a platoon shown in Fig.~\ref{fig_DeadlockBreaking}, the planned trajectories of these two vehicles can move in parallel and maintain a constant distance to the reference trajectory. We consider this situation as a deadlock since they cannot travel at their target lanes.

Based on this observation, we propose the following criterion to decide whether vehicle $i$ is stuck in a deadlock:
\begin{equation}
\begin{aligned}
\left| \max\{d(\mathbf{x}_i^{-n},\mathbf{x}_i^{ref})\} - \min\{d(\mathbf{x}_i^{-n},\mathbf{x}_i^{ref})\} \right|  &\leq \epsilon_1 \  \land \\
\left| \rm{mean}\{d(\mathbf{x}_i^{-n},\mathbf{x}_i^{ref})\}  \right| &\geq \epsilon_2 ,
\end{aligned}
\label{Eq_Deadlock}
\end{equation}
where $\mathbf{x}_i^{-n} = \left[ x_i^{H-n+1} ; x_i^{H-n+2} ; ... ; x_i^H\right]$ is the last $n$ points of the planned trajectory, $d(\mathbf{x}_i^{-n},\mathbf{x}_i^{ref}) \in \mathbb{R}^n$ is the distance from these points to the reference trajectory, and $\epsilon_1$ and $\epsilon_2$ are tunable thresholds.

To break the deadlock, one solution is to change the desired speeds of these vehicles, which will result in different reference trajectories. We assign the following rules. The front vehicle has the highest priority. If there are multiple vehicles moving in parallel, we specify the one who has smaller $\rm{mean}\{d(\mathbf{x}_i^{-n},\mathbf{x}_i^{ref})\}$ will be assigned a larger speed. Finally, If two vehicles are symmetric about the reference trajectory, as shown in Fig. \ref{fig_Deadlock}, the vehicle merging from the left will have higher priority. For example, in Fig. \ref{fig_DeadlockBreaking}, vehicle 1 is assigned a larger speed than vehicle 2, and the deadlock is resolved. Once they converge to the reference trajectory, they can move at their original desired speed.

This strategy of deadlock breaking assigns different speeds to vehicles and prevents vehicles from having identical or symmetric reference trajectories, as shown in Fig. \ref{fig_Deadlock}. Therefore, this strategy breaks the symmetry between vehicles that leads to a deadlock. Compared to the priority constraints proposed in MCCFS algorithm \cite{MCCFS2020} for deadlock resolution, our method do not increase the number of constraints. In the optimization problem \eqref{Eq_DCFS_opt}, only the reference trajectory $\mathbf{x}_i^{ref}$ is changed by changing the desired speed and the constraints are not affected.

We define the \textbf{consensus} in this paper as the situation when all vehicles plan trajectories to reach their reference trajectories without inter-vehicle collision.

\subsection{System Architecture}
\begin{figure}
	\centering
	\includegraphics[width=0.5\textwidth]{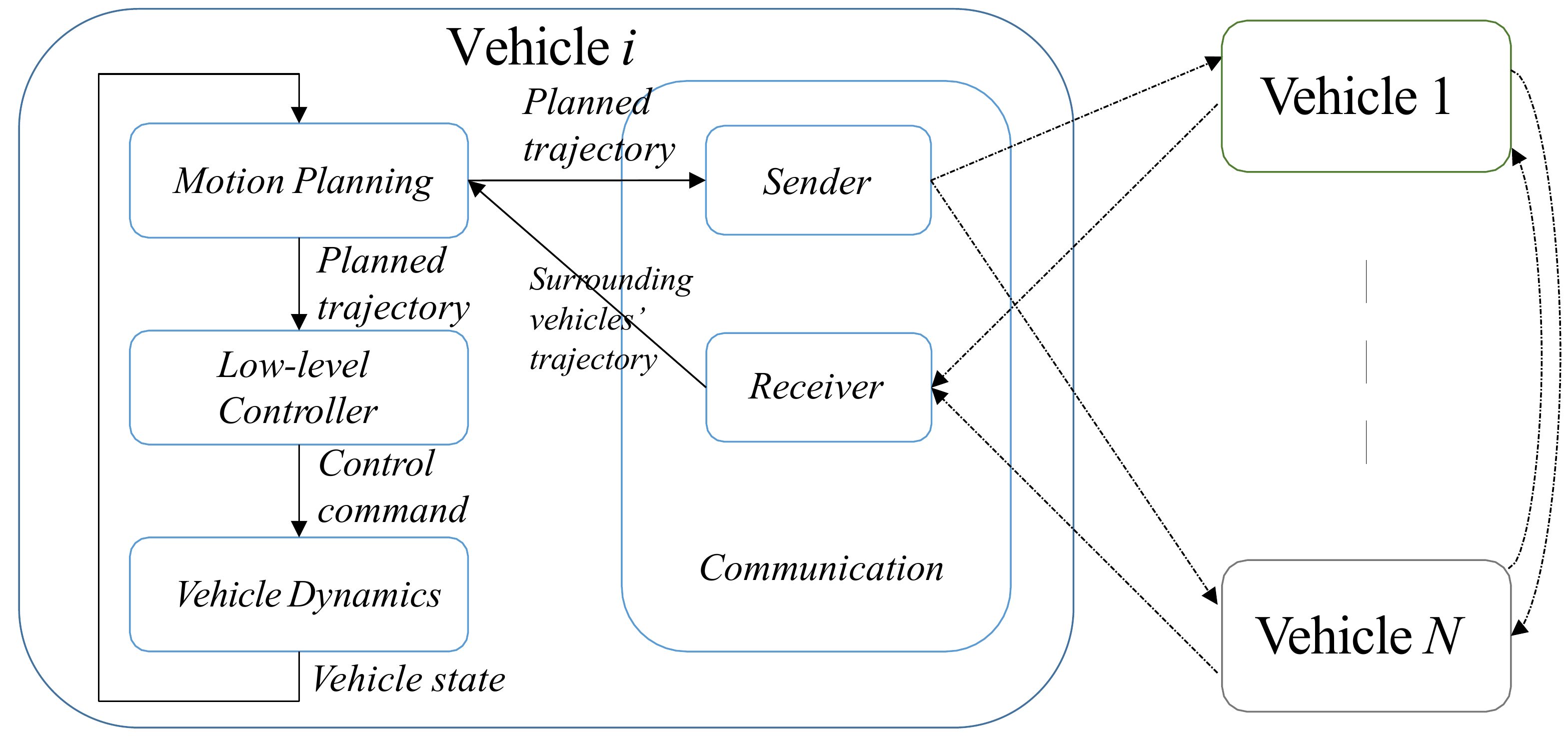}
	\vspace{-7mm}		
	\caption{System architecture.}
	\label{fig_SysArch}
	\vspace{-5mm}
\end{figure}

The system architecture with respect to the communication, planning, and control scheme is shown in Fig. \ref{fig_SysArch}. First, trajectories are shared among vehicles. Based on the trajectory information, its current state, and the reference trajectory, the vehicle plans an optimal trajectory. The low-level controller then computes the control command to follow the planned trajectory. Then the control command is executed by the vehicle which then returns updated vehicle states. This process iterates every replanning time $T_r$ and ensures the system is robust to tracking errors.

\section{Numerical Results}
\label{Ch4}

This section presents the simulation results of the proposed method on six driving scenarios: unstructured road, intersection, crossing, platoon formation, merging, and overtaking. The CFS-DMPC design is compared with the centralized method MCCFS \cite{MCCFS2020} and RVO \cite{van2008reciprocal}. All methods are tested on a laptop with 2.60GHz Intel Core i7-9750H in Python script. Section \ref{subsec_SimSetup} describes the simulation setup. Section \ref{subsec_SimResults} presents the simulation results. Section \ref{subsec_Performance} analyzes the performance of the proposed method. 

\subsection{Simulation Setup}
\label{subsec_SimSetup}

\subsubsection{Road Environment}
\begin{figure}
	\centering
	\includegraphics[width=0.45\textwidth]{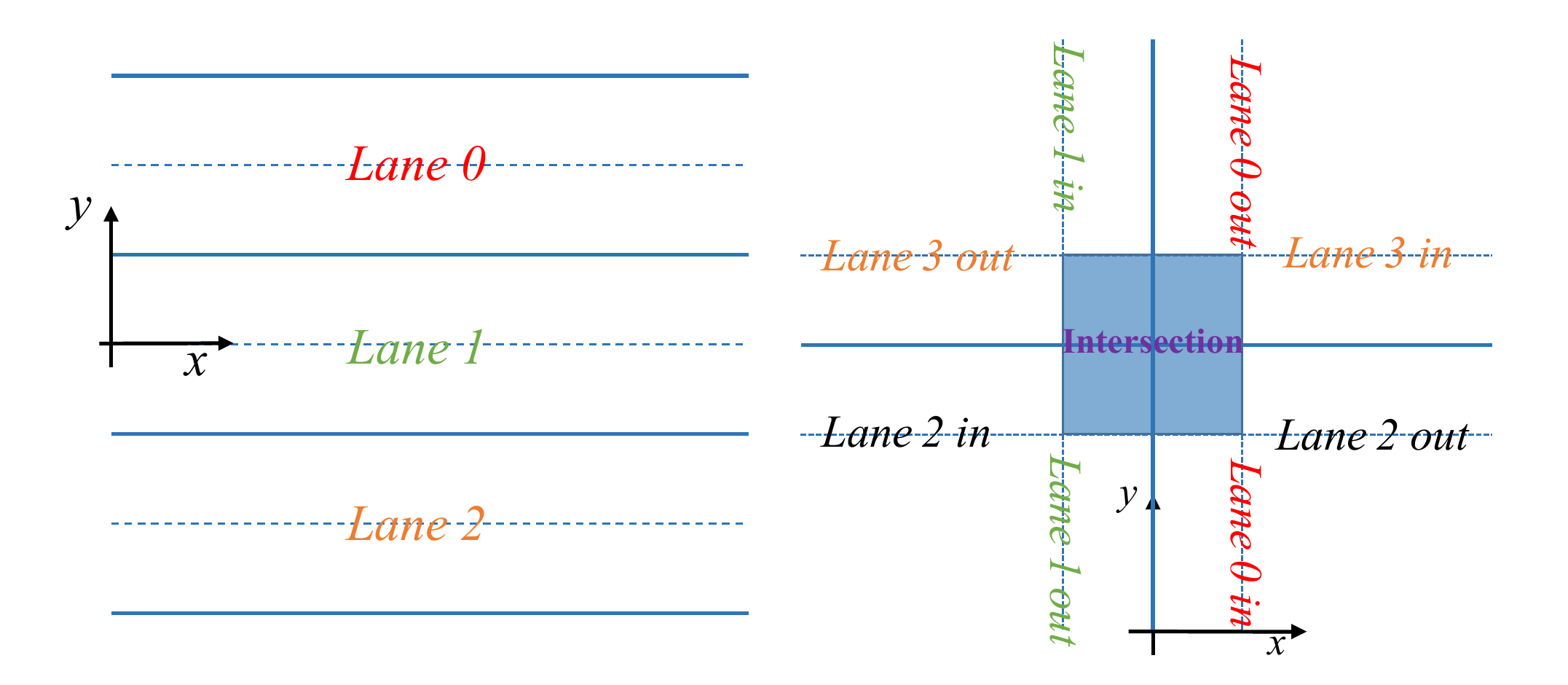}
	\vspace{-4mm}		
	\caption{The road environment. Left: highway. Right: intersection.}
	\label{fig_Road}
\end{figure}

The road environments are shown in Fig. \ref{fig_Road}. The solid line is the boundary of lane, the dash line is the centerline, and the lane width is 4$m$. There are three lanes in the highway environment. Note that the $x$-axis aligns with the centerline of $Lane\ 1$. In the intersection environment, there are four lanes, each with an incoming part and an outgoing part. The $2m \times 2m$ shaded block is called the intersection area. The $y$-axis aligns with the boundary of $Lane\ 0$ and $Lane\ 1$.

%

\begin{figure}[t]
	\centering
	\subfigure[The planned trajectories at $t=24T_r$. The triangle represents the starting position of the planned trajectory, the dash line represents the circle diameter.]{	\includegraphics[width=0.49\textwidth]{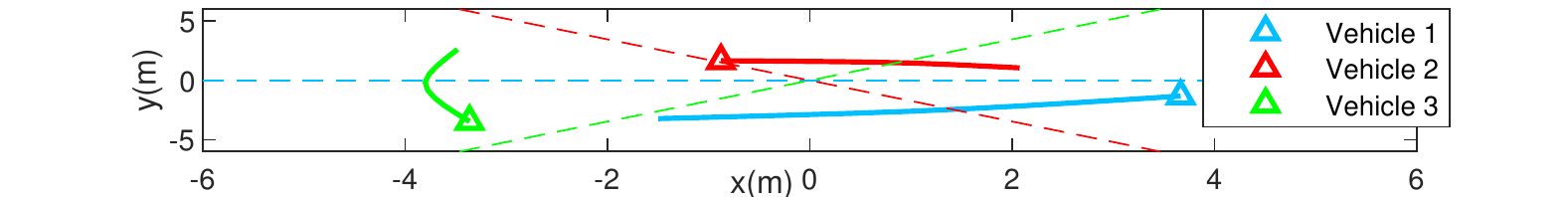}\label{fig_Results_Intersection_20}}
	\subfigure[The simulation result. The circle represents the vehicle's initial position, the square represents the vehicle's end position, and the star marks the position every $5T_r$.]{\includegraphics[width=0.49\textwidth]{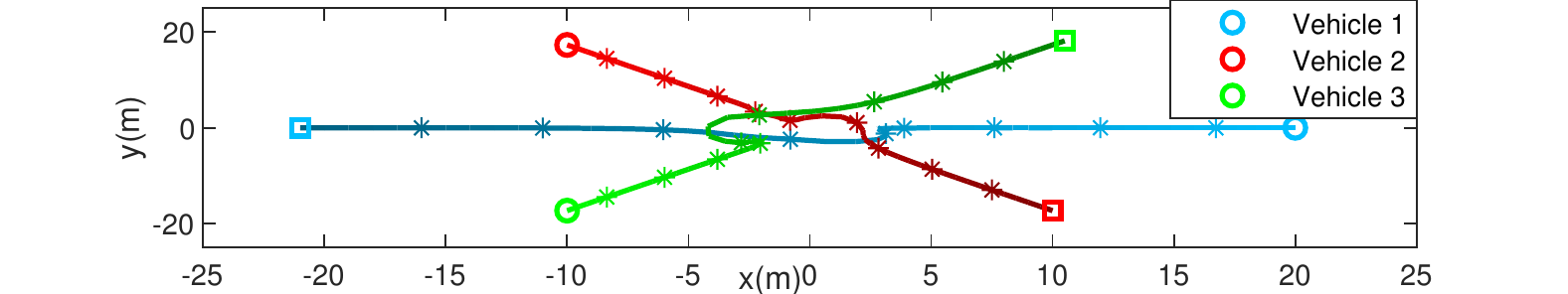}\label{fig_Results_UnstructedRoad_all}}
	\vspace{-5mm}	
	\caption{Unstructured road situation.}
	\label{fig_Results_UnstructedRoad}
	\vspace{-5mm}
\end{figure}

\begin{figure}[t]
	\centering
	\subfigure[The planned trajectories at $t=20T_r$. The triangle represents the starting position of the planned trajectory, the cross marks the planned trajectory every $3T_s$, the dash line represents the lane centerline.]{	\includegraphics[width=0.49\textwidth]{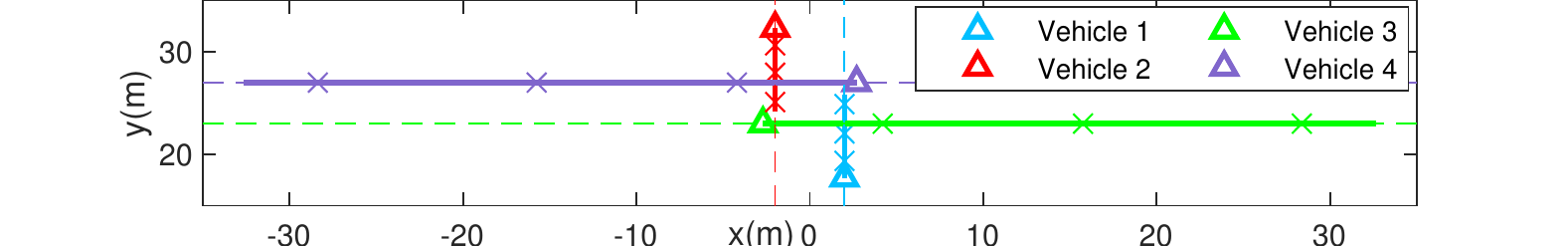}\label{fig_Results_Intersection_20}}
	\subfigure[The simulation result. The circle represents the vehicle's initial position, the square represents the vehicle's end position, and the star marks the position every $5T_r$.]{	\includegraphics[width=0.49\textwidth]{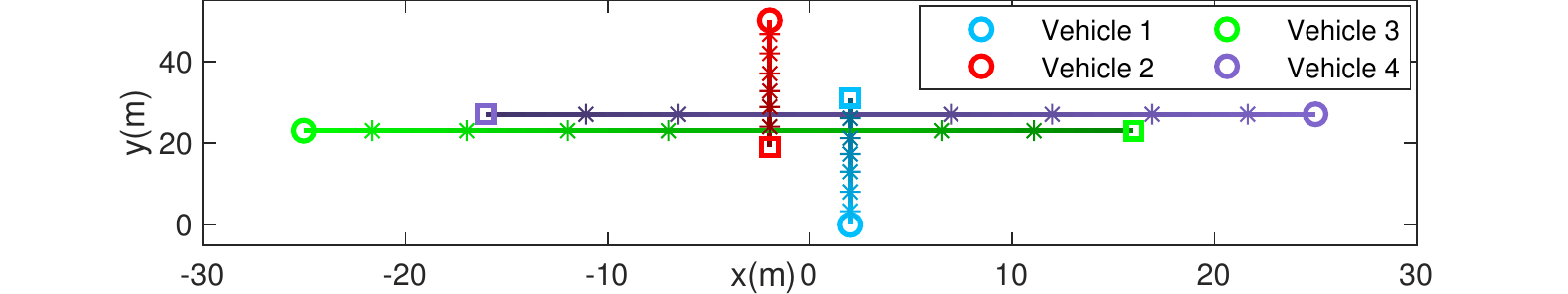}\label{fig_Results_Intersection_all}}
	\vspace{-7mm}		
	\caption{Intersection situation.}
	\label{fig_Results_Intersection}
\end{figure}

\begin{figure}[t]
	\centering
	\includegraphics[width=0.49\textwidth]{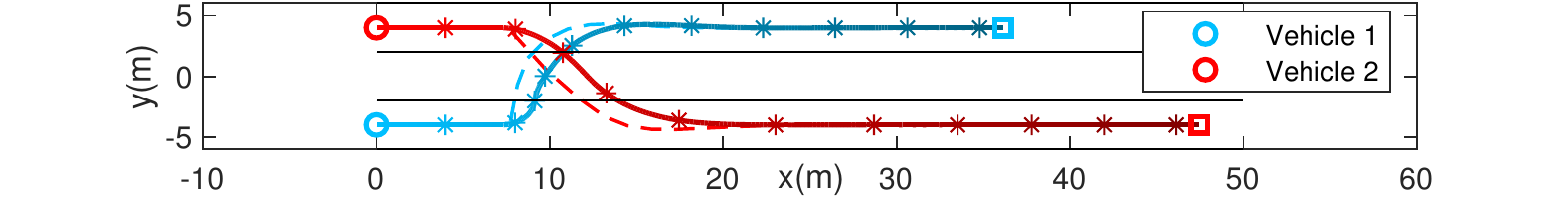}
	\vspace{-8mm}		
	\caption{Crossing situation. The circle represents the vehicle's initial position, the square represents the vehicle's end position, and the star marks the vehicle's position every $25T_r$. The dash line is the planned trajectory when vehicles reach a consensus.}
	\label{fig_Results_Crossing}
	\vspace{-5mm}
\end{figure} 

\begin{figure}[t]
	\centering
	\subfigure[The planned trajectories (dash lines) every $5T_r$ and the executed trajectory.]{	\includegraphics[width=0.49\textwidth]{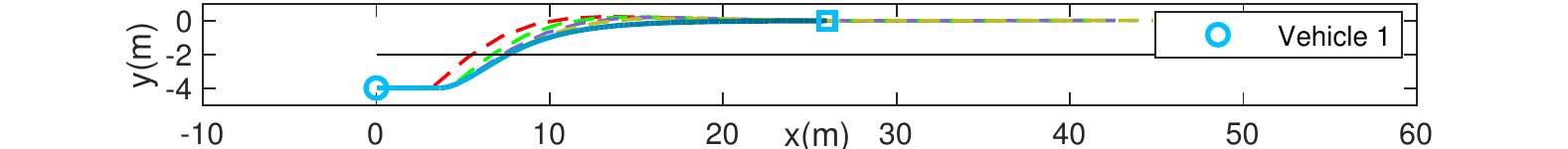}\label{fig_Results_Platoon_vehicle1}}
	\subfigure[The simulation result. The dash line is the planned trajectory when vehicles reach a consensus.]{	\includegraphics[width=0.49\textwidth]{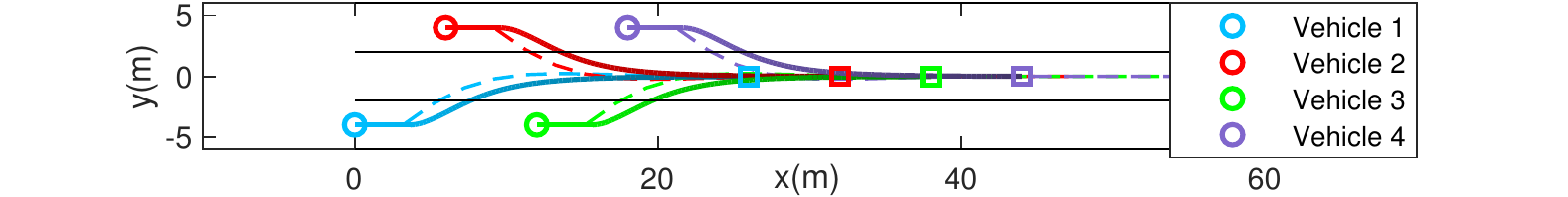}\label{fig_Results_Platoon_all}}
	\vspace{-4mm}		
	\caption{Platoon formation. The circle represents the vehicle's initial position and the square represents the vehicle's end position. }
	\label{fig_Results_Platoon}
\end{figure}

\begin{figure}[t]
	\centering
	\includegraphics[width=0.49\textwidth]{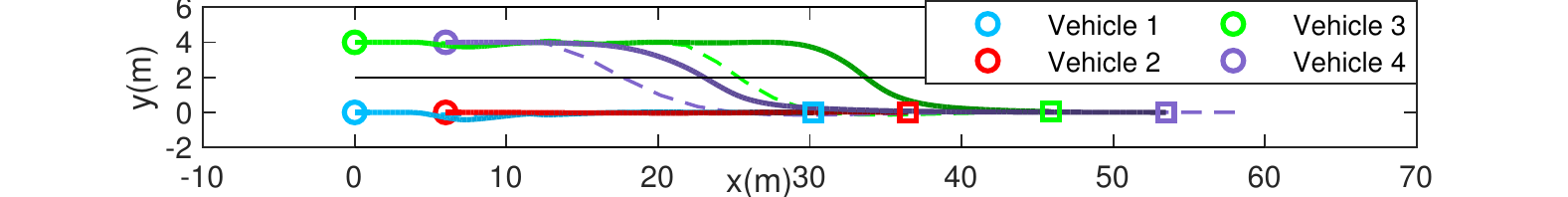}
	\vspace{-7mm}		
	\caption{Merging situation. The circle represents the vehicle's initial position and the square represents the vehicle's end position. The dash line is the planned trajectory when vehicles reach a consensus.}
	\label{fig_Results_Merging}
	\vspace{-5mm}
\end{figure} 

\begin{figure}[t]
	\centering
	\includegraphics[width=0.49\textwidth]{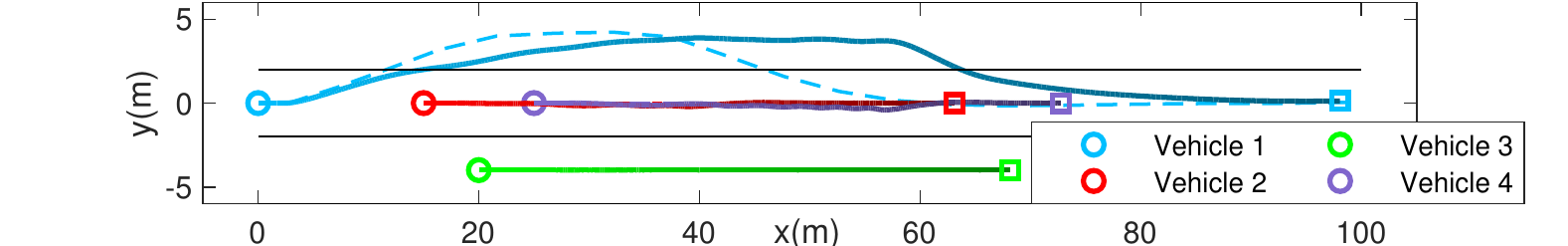}
	\vspace{-7mm}	
	\caption{Overtaking situation. The circle represents the vehicle's initial position and the square represents the vehicle's end position. The dash line is the planned trajectory when vehicles reach a consensus.}
	\label{fig_Results_Overtaking}
\end{figure}

\subsubsection{Vehicle Model}
The vehicle bicycle kinematic model is applied for vehicle modeling. A vehicle's state includes the position $(x_0,y_0)$, the velocity $v_0$, and the heading $\theta_0$. The control input are the acceleration $a$ and the steering angle $\delta$. $L$ is the wheel base. Assuming that the steering angle $\delta$ is constant during a replanning time $T_r$, the vehicle rotates around an instant center $O$ with a rotation radius $R$. In one replanning time $T_r$, the distance that the vehicle travels is $L_r = v_0 T_r+ \frac{1}{2} a T_r^2$ and the curvature is $\kappa = \frac{\tan \delta}{L_r}$. Then the state is updated as follows: $v_1 = v_o + a T_r$, $\theta_1 = \theta_o + \int_{0}^{L_r} \kappa \mathrm{d}s =  \theta_o + \kappa L_r$, $	x_1 = x_o + \int_{0}^{L_r} \cos(\theta_o + \kappa s) \mathrm{d}s =  x_o + \frac{\sin(\theta_o + \kappa L_r)-\sin(\theta_0)}{\kappa}$, and $y_1 = y_o + \int_{0}^{L_r} \sin(\theta_o + \kappa s) \mathrm{d}s =  y_o + \frac{\cos(\theta_0)-\cos(\theta_o + \kappa L_r)}{\kappa}$.



\subsubsection{Low-level Controller}
We use a low-level tracking controller proposed in \cite{chen2018foad}. The control input $a$ $\in$ $[-5, 5]\ m/s^2$ and $\delta$ $\in$ $[-45^{\circ}, 45^{\circ}]\ $ are functions of the difference to the planned trajectory, the desired speed, and the desired angle with respect to the planned trajectory. They remain constant in a replanning time $T_r$. 

\subsection{Simulation Results}
\label{subsec_SimResults}

For vehicle representation, we choose $r = 3$, $l = 1.9$, and $w = 1$, considering the vehicle size in real world. In the intersection secnario, $r$ is changed to $2.5$. In the unstructured road and intersection scenario, we only consider trajectory planning and assume the vehicle can reach the planned position exactly at the next time step. Therefore, the replanning time $T_r = T_s$. In other scenarios, we choose $T_r=0.02s$ and use the low-level controller. The reference trajectory of each vehicle is set as the centerline of its target lane.

\subsubsection{Unstructured Road Scenario}
In this scenario, three vehicles are symmetrically located on a circle with radius $20m$. The reference trajectories are straight lines along the diameters with speed $10m/s$. We let vehicles travel to the opposite points on the circle. The planning horizon is $H = 10$ and the sampling time is $T_s = 0.1s$. Figure \ref{fig_Results_UnstructedRoad} shows the results from the proposed method, which finds a safe coordination strategy that is to rotate around point $(0,0)$ at $t=24T_r$. Vehicles reach this consensus in $6T_r$.

\subsubsection{Intersection Scenario}
In this scenario, four vehicles travel at four-way-stop intersection with speed $10m/s$ and we add constraints disallowing them to deviate from the centerline. Their initial positions are $(2,0)$, $(-2,50)$, $(-25,23)$, and $(25,27)$, respectively. We choose $H = 10$ and $T_s = 0.1s$. Instead of using $\mathbf{x}_i^{ref}$ in \eqref{Eq_Deadlock} to break the deadlock, we measure the distance to the point where an outgoing lane intersects with the intersection area. We choose $n = 2$, $\epsilon_1 = 0.15$, and $\epsilon_2 = 2$. The result is given in Fig. \ref{fig_Results_Intersection}. At $t=20T_r$, vehicles 3 and 4 first trigger deadlock resolution and change their speeds. The planned trajectories show the consensus among vehicles, such that vehicles 3 and 4 pass the intersection before other two vehicles.



\subsubsection{Crossing Scenario}
In this scenario, two vehicles with desired speed $10m/s$ start at $(0,-4)$ and $(0,4)$ respectively. Vehicle 1 moves from $Lane\ 2$ to $Lane\ 0$, and vehicle 2 from $Lane\ 0$ to $Lane\ 2$. We choose $H = 20$, $T_s = 0.1s$, $n = 5$, $\epsilon_1 = 0.01$, and $\epsilon_2 = 0.2$. The result is shown in Fig. \ref{fig_Results_Crossing}. When changing lane, two vehicles both plan trajectories that converge to $Lane\ 1$, which triggers the deadlock resolution. Vehicle 2 speeds up and vehicle 1 maintains its speed, such that vehicle 2 crosses first. The consensus is reached in $4T_r$.


\subsubsection{Platoon Formation}
In this scenario, four vehicles traveling at either $Lane\ 0$ or $Lane\ 2$ with desired speed $20 m/s$ are forming a platoon at $Lane\ 1$. Their initial positions are $(0,-4)$, $(6,4)$, $(12,-4)$, and $(18,4)$, respectively. We choose $H = 20$ and $T_s = 0.1s$. The result is given in Fig. \ref{fig_Results_Platoon}. Four vehicles successfully form a platoon and maintain safe distance with each other. The consensus is reached in $1 T_r$.


\subsubsection{Merging Scenario}
Two vehicles travels at $Lane\ 1$, and another two vehicles at $Lane\ 0$ intend to merge into $Lane\ 1$, all with speed of $10 m/s$. We choose $H = 25$, $T_s = 0.1s$, $n = 5$, $\epsilon_1 = 0.01$, and $\epsilon_2 = 0.2$. The result is presented in Fig. \ref{fig_Results_Merging}. Unlike the platoon formation scenario, four vehicles can be stuck in a deadlock and travel in parallel. When a deadlock is detected, vehicle 4 changes its speed to $25 m/s$ and vehicle 3 speeds up to $20 m/s$. Vehicles 1 and 2 maintain their speeds since they are moving on their target lane. Then vehicles 4 and 3 merge into $Lane\ 1$ in sequence in front of vehicles 1 and 2. Vehicles use $3T_r$ to reach the consensus.


\subsubsection{Overtaking Scenario}
In this scenario, vehicle 1 traveling at $50 m/s$ overtakes other three vehicles traveling at $10 m/s$. Their initial positions are $(0,0)$, $(15,0)$, $(20,-4)$ and $(25,0)$, respectively. We choose $H = 25$ and $T_s = 0.1s$. The result is shown in Fig. \ref{fig_Results_Overtaking}. Vehicle 1 changes lane in order to avoid collision with vehicles being overtaken, and other vehicles can follow their reference trajectories. The consensus among vehicles is reached in $9T_r$.


\subsection{Performance Analysis}
\label{subsec_Performance}
\subsubsection{Distributed Versus Centralized Design}
The proposed CFS-DMPC design is compared to a centralized approach, MCCFS \cite{MCCFS2020}, in terms of computation time and the total cost. We implement MCCFS in a MPC framework and allow it to converge at every time step. Note that the total cost can be negative since $\frac{1}{2}c_o(\mathbf{x}_i^{ref}) ^\top\mathbf{x}_i^{ref}$ is neglected. We simulate the formation of two to five vehicles. We choose $H=20$, $T_r = 0.02s$, $T_s = 0.1s$, and total simulation time $t = 100T_r$ in the cases with tracking control. The parameters are $H=20$, $T_r = T_s = 0.1s$, and $t = 30T_r$ for the cases without a controller. The result is given in Table \ref{table_preformance} and Table \ref{table_opt}. 

Both methods ensure safe motion coordination. However, Table \ref{table_preformance} shows that the CFS-DMPC design outperforms the MCCFS by more than an order of magnitude in terms of total average and maximum time. Besides, since each vehicle only solves a QP problem in the CFS-DMPC design, the computation time is short and suitable for online implementation. In Table \ref{table_opt} (the first and second multicolumn), the MCCFS achieves smaller total costs. The results in Table \ref{table_preformance} and Table \ref{table_opt} show the trade-off between computation time and the optimality of the planned trajectory.

\renewcommand{\arraystretch}{1.5} 
\begin{table}[t]
	\centering
	\scriptsize
	\begin{threeparttable}
		\caption{Computation time (in second) for centralized and distributed approaches.}
		\label{table_preformance}
		\begin{tabular}{cccccc}
			\toprule
			
			No. & \multicolumn{2}{c}{Centralized} & \multicolumn{3}{c}{Distributed}		\cr
			\cmidrule(lr){2-3} \cmidrule(lr){4-6} & Avg. & Max. & Avg. (Each) &   Avg. (Total) &  Max. (Total)  \cr
			
			\midrule
			
			2 & 0.1853 & 0.2293 & 0.0048 & 0.0096 & 0.0124  \cr
			3 & 0.4313 & 0.4726 & 0.0091 & 0.0272 & 0.0381  \cr
			4 & 0.7768 & 1.1257 & 0.0129 & 0.0514 & 0.0700  \cr
			5 & 1.2780 & 1.7829 & 0.0183 & 0.0913 & 0.1667  \cr
			
			\bottomrule
		\end{tabular}
	\end{threeparttable}
	\vspace{-6mm}
\end{table}

\renewcommand{\arraystretch}{1.5} 
\begin{table}[t]
	\centering
	\scriptsize
	\begin{threeparttable}
		\caption{The total cost for both  centralized and distributed approaches with and without tracking errors.}
		\label{table_opt}
		\begin{tabular}{ccccccccc}
			\toprule
			
			No. & \multicolumn{2}{c}{w/} & \multicolumn{2}{c}{w/o}	& \multicolumn{3}{c}{Distr.}	\cr
			\cmidrule(lr){2-3} \cmidrule(lr){4-5} \cmidrule(lr){6-8} & Centr. & Distr. & Centr. & Distr. & w/ & w/o & Ratio ($\frac{\text{w/}}{\text{w/o}}$) \cr
			
			\midrule
			
			2 & -19.41 & -16.32  & -18.25 & -9.13 & -11.22 & -18.12 & 61.93\% \cr
			3 & -34.99 & -28.21  & -30.35 & -15.17 & -25.82 & -37.98 & 67.99\% \cr
			4 & -55.44 & -43.06  & -44.71 & -22.34 & -49.30 & -67.92 & 72.58\%   \cr
			5 & -81.49 & -61.23  & -61.56 & -30.75 & -83.80 & -110.10 & 76.12\%   \cr

			\bottomrule
		\end{tabular}
	\end{threeparttable}
\end{table}

%
%
%
%
%
%

\subsubsection{Tracking Errors}
The vehicle dynamics and control input saturation are not taken into account in the CFS-DMPC design. As a result, the planned trajectory may violate these constraints and the vehicle cannot track the planned trajectory perfectly. The tracking errors jeopardize the optimality of the executed trajectory. Taking platoon formation as an example, the difference between planned trajectories and the executed trajectory of vehicle 1 is presented in Fig. \ref{fig_Results_Platoon_vehicle1}. During lane changing, vehicle 1 has a mean cross-track error of $0.023m$. It tends to react slowly and deviates from the optimal trajectory, which takes the vehicle more time and distance to converge to the target lane. The total cost with and without tracking errors is presented in Table \ref{table_opt} (the third multicolumn). Relevant parameters are $H=20$, $T_r = T_s = 0.02s$, and $t = 60T_r$. The tracking errors result in loss of around $25-40\%$ optimality. In the presence of tracking errors, the proposed method is able to safely and efficiently coordinate multiple vehicles in various scenarios, which demonstrate its robustness.

%
%
%
%

\subsubsection{ Integrated design Versus Decoupled scheduling and planning}
\cite{liu2017distributed} uses a decision maker to decide passing order and a speed profile planner in intersection scenario. The optimality and computational complexity rely on conflict zone resolution. In contrary, our method does not decouple decision making and motion planning, and does not formulate conflict zones explicitly. The proposed method take into account the interaction among vehicles and allow vehicles to reach a consensus by V2V communication. Therefore, a decision maker for a scheduling is not necessary in our system. Our method relies on a simple deadlock resolution in a four-vehicle intersection case, but its robustness has not been tested yet when traffic density goes up.

\subsubsection{CFS-DMPC Versus RVO}
We simulate the unstructured road scenario to compare the CFS-DMPC design with RVO, in terms of average trajectory length of each vehicle, time duration to reach goal, and computation time. Note that vehicle dynamics is not considered. The result is summarized in Table \ref{table_RVO_MPC}. The performance of our method is comparable to RVO in terms of trajectory length. The computation time of RVO is two to three times faster, while the CFS-DMPC design is more optimal than RVO in time duration. However, with increasing number of agents (e.g. larger than 10), fine tuning for CFS-DMPC is required to obtain a good result. Therefore, the scalability in such tight environments may be an issue.

\renewcommand{\arraystretch}{1.5} 
\begin{table}[t]
	\centering
	\scriptsize
	\begin{threeparttable}
		\caption{Comparison between RVO and CFS-DMPC.}
		\label{table_RVO_MPC}
		\begin{tabular}{ccccccc}
			\toprule
			
			No. & \multicolumn{2}{c}{Avg. Length (m)} & \multicolumn{2}{c}{Time Duration (s)} & \multicolumn{2}{c}{Avg. Computation Time (s)}		\cr
			\cmidrule(lr){2-3} \cmidrule(lr){4-5}  \cmidrule(lr){6-7} & RVO & MPC & RVO & MPC & RVO (Each) & MPC (Each) \cr
			
			\midrule
			2 & 42.57 & 41.61 & 5.5 & 5.3 & 0.0015 & 0.0022   \cr
			4 & 44.85 & 48.63 & 8.2 & 6.4 & 0.0022 & 0.0052   \cr			
			6 & 43.97 & 45.85 & 7.2 & 5.7 & 0.0027 & 0.0083   \cr
			
			\bottomrule
		\end{tabular}
	\end{threeparttable}
	\vspace{-6mm}
\end{table}

%
%
%
%
%

\section{Conclusions}
\label{Ch5}

This paper proposed a CFS-based distributed MPC (CFS-DMPC) approach for multi-vehicle motion coordination. Assuming that vehicles are able to communicate with each other, the coupling in the centralized problem is removed to formulate a distributed MPC. Using CFS algorithm, the collision avoidance constraints are convexified and the sub-problem is transformed into a QP problem, which can be solved efficiently. We showed how to detect a deadlock and resolve it by changing vehicles' desired speeds. The proposed approach was validated through numerical simulations. We showed that the CFS-DMPC design outperforms MCCFS by more than an order of magnitude in computation time. We showed that our method is robust to tracking errors and that a decision maker for a passing order is not necessary in our system.

For future work, we will test the CFS-DMPC with real
vehicles considering various challenges, e.g., realistic vehicle dynamics, tracking errors caused by disturbances, communication delays, misaligned sampling time, etc. The CFS-DMPC can consider realistic vehicle dynamics and disturbances to reduce tracking errors. Besides, an asynchronous planning mechanism can be easily developed since all communicated information contains time stamps.  Lastly, while the proposed method results in converging trajectories in all simulations, we plan to formally analyze its theoretical stability and robustness.

\addtolength{\textheight}{-12cm}   



%


\bibliographystyle{IEEEtran}
\bibliography{Reference.bib}

\end{document}